\documentclass{llncs}
\usepackage{epsf}

\usepackage[english]{babel}
\usepackage[utf8x]{inputenc}
\usepackage{geometry}

\usepackage{amsmath}
\usepackage{graphicx}
\usepackage{float}
\usepackage[justification=justified]{caption}
\usepackage{enumitem} 
\usepackage{dirtytalk}
\usepackage{array}

\usepackage{hyperref} 
\usepackage{breakurl}

\usepackage{rotating}
\usepackage{placeins}
\usepackage{multirow}

\usepackage{xcolor}

\newcommand{\citet}{\cite}
\newcommand{\citep}{\cite}

\begin{document}

\title{Adaptation of Machine Translation Models with Back-translated Data using Transductive Data Selection Methods}
       
\author{Alberto Poncelas, Gideon Maillette de Buy Wenniger, Andy Way}

\institute{
    ADAPT Centre, School of Computing, \\ 
    Dublin City University, Dublin, Ireland\\
    {\tt \{firstname.lastname\}@adaptcentre.ie}
}

\maketitle
\begin{abstract}
Data selection has proven its merit for improving Neural Machine Translation (NMT), when applied to authentic data. But the benefit of using synthetic data in NMT training, produced by the popular back-translation technique, raises the question if data selection could also be useful for synthetic data?

In this work we use Infrequent {\em n}-gram Recovery (INR) and Feature Decay Algorithms (FDA), two transductive data selection methods to obtain subsets of sentences from synthetic data. These methods ensure that selected sentences share {\em n}-grams with the test set so the NMT model can be adapted to translate it.

Performing data selection on back-translated data creates new challenges as the source-side may contain noise originated by the model used in the back-translation. Hence, finding {\em n}-grams present in the test set become more difficult. Despite that, in our work we show that adapting a model with a selection of synthetic data is an useful approach.


\end{abstract}

\section{Introduction}

Neural Machine Translation (NMT) models tend to perform better with larger amounts of data. However, a smaller model trained with data in the same domain as the document to be translated (test set) may perform better than a bigger general-domain model.

Data selection algorithms can be applied as a technique to obtain data of a particular domain. Generally speaking, these methods start from a large set of sentences, and from this set select a subset of sentences that are closer to the domain of interest than other sentences in the large set. Among these methods, Transductive Algorithms (TA) perform the selection by using the test set as seed and retrieving those sentences that are relatively closer to this seed than others. Models built using the output of TA also perform better than general-domain models \citep{poncelas2018feature,silva2018extracting}. 

Alternatively, a general-domain model can also be adapted to a certain domain by applying the technique known as \textit{fine-tuning} \citep{luong2015stanford,freitag2016fast,van2017dynamic}. This consists of training the last epochs of an NMT model (built with out-domain data) using a smaller but in-domain set of sentences.

Unfortunately, additional data that are closer to the test set are not always available. The work of \citet{sennrich2015improving} showed that the inclusion of back-translated data can boost the performance of NMT models. Since then, adding synthetic data for training Machine Translation (MT) models has become more popular.

\begin{figure*}[hbt]
\includegraphics[width=12cm, height=7cm]{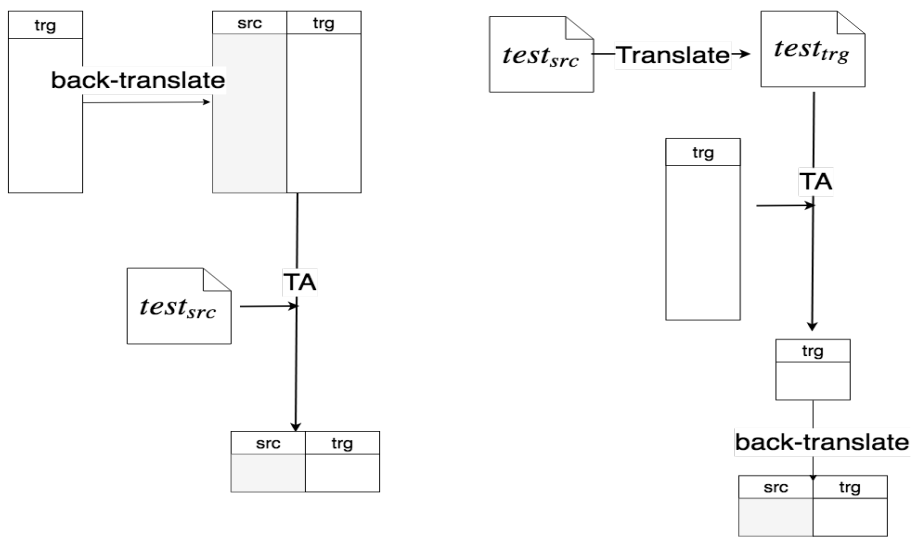}
\centering
\caption{Pipeline of the batch (left) and online (right) processing to obtain TA-selected synthetic data.
\label{fig:batch_vs_online}}
\end{figure*}

In this work we want to investigate whether it is useful to apply TA to synthetic data selection, in order to retrieve artificial sentences closer to the test set. We study the performance of TA on the task of synthetic data selection, applied in two different configurations (see Figure \ref{fig:batch_vs_online}):

\begin{enumerate}
\item Batch processing: The first approach involves back-translating a monolingual set of sentences completely and then selecting sentences from synthetic parallel set. The selection criteria of TA are based on the overlap of {\em n}-grams of the test set (the seed) with those in the source-side of the parallel set. For this reason, the performance of TA may be worse on back-translated data as the {\em n}-grams, which have been artificially generated, may be unnatural in terms of word-order.
\item Online processing: This involves selecting the necessary monolingual, target-side, sentences and afterwards back-translating the selected set. The advantage of the online process is that it is not necessary to back-translate the complete data set before selecting data. Nevertheless, as the selection is performed in monolingual target-language we cannot use the test set (which is in the source-side language) as seed. To solve this, we can proceed as described in the work of \citet{poncelas2018data} and translate the test set using a generic-domain NMT model. Then, this translated text can be used as seed. 
\end{enumerate}

\section{Related Work}

\subsection{Transductive Data Selection Algorithms}
\label{sec:transductive_DSA}

In this section we describe the algorithms used in the paper, which belong to the family of transductive \citep{Vapnik1998} data selection methods. Such methods select the most relevant sentences for the test set using the (source-side) test set itself. The methods score each sentence $s$ in the candidate data $U$ (the set of sentences that have not been yet selected), and then the sentence with the highest score is added to selected pool $L$, which is initially empty. Note that this process is done iteratively as the scores (which depend on $U$ and $L$) are updated after a sentence has been selected.

\paragraph{Infrequent {\em n}-gram Recovery} (INR):
In the work of \citet{parcheta2018data,gasco2012does} they propose extracting sentences containing {\em n}-grams (present in the test set) that are considered infrequent. Therefore, words such as stop words are ignored. The sentences in the candidate data $U$ are scored according to Equation \eqref{eq:infreq_ngr_recover}:

\begin{equation}\label{eq:infreq_ngr_recover}
score(s,U)=\sum_{ngr \in S_{test}}  max(0,t-C_{S_I+L}(ngr)) 
\end{equation}

\noindent where $t$ is the threshold that indicates whether an {\em n}-gram is frequent or not. If the count of the {\em n}-gram $ngr$ ($C_{S_I+L}(ngr)$) in the selected pool $L$ (and an in-domain set $S_I$ used for initialization) exceeds the value of $t$ then it will not contribute to the score of the sentence.

\paragraph{Feature Decay Algorithms} (FDA):
Feature Decay Algorithms \citet{biccici2011instance} selects data by promoting sentences containing many {\em n}-grams from the test set, but penalizing those {\em n}-grams that have been selected several times. Each {\em n}-gram $ngr$ is assigned an initial score, then each time a sentence containing $ngr$ is selected the score of $ngr$ is decreased. The default scoring function is defined as in Equation \eqref{eq:fda_sentencescore}:

\begin{equation}\label{eq:fda_sentencescore}
score(s,L)=\frac{\sum_{ngr \in S_{test}} 0.5^{C_L(ngr)}}{length(s)}
\end{equation}

Observe that the more occurrences of $ngr$ are in the selected pool $L$ ($C_L(ngr)$) the less it contributes towards the scoring of the sentence $s$.

\subsection{Using Approximated Target Side}
\label{sec:approximated_target_side}

The methods presented in Section \ref{sec:transductive_DSA} use the test set as seed in order to retrieve sentences. However, a similar approach can be executed by using an approximated translation of the test set (approximated target side) as seed \citep{poncelas2018data}. This seed can be generated by another MT model.

The output of a TA, such as INR or FDA, can be represented as a sequence of sentences $TA_{src}=(s_1^{(src)},s_2^{(src)},s_3^{(src)},...s_{N}^{(src)})$ of $N$ sentences. We use the subscript \textit{src} to indicate that the seed is a text in the source language. However, we can first translate the test set using a generic NMT model and execute the TA using the translation as a seed. The output of this execution could also be represented as a sequence of sentences $TA_{trg}=(s_1^{(trg)},s_2^{(trg)},s_3^{(trg)},...s_{N}^{(trg)})$


The two outputs, $TA_{src}$ and $TA_{trg}$, can be combined as a new sequence of $N$ sentences as in Equation \eqref{eq:FDA_comb_seq}

\begin{equation}\label{eq:FDA_comb_seq}
TA=(s_1^{(src)},...s_{N*\alpha}^{(src)},s_1^{(trg)},...s_{N* (1-\alpha)}^{(trg)})
\end{equation}


\noindent where the top sentences from each output are concatenated. The value of $\alpha \in [0,1]$ represents the proportion of data that are selected from $TA_{src}$ and $TA_{trg}$.

\begin{figure*}[hbt]
\includegraphics[width=12cm, height=8cm]{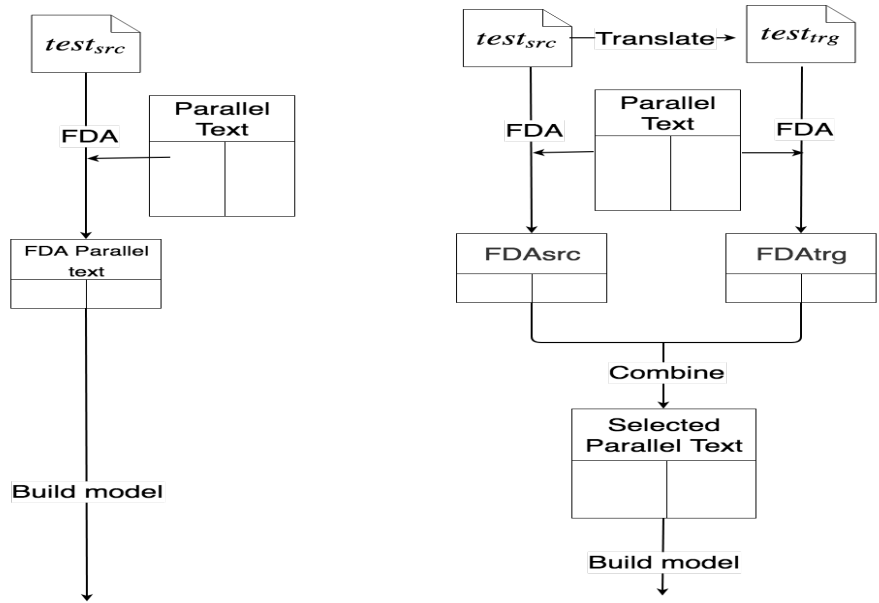}
\centering
\caption{Pipeline of the traditional usage of FDA (left) and pipeline of our proposal, using the target-side (right) \citep{poncelas2018data}.
\label{src_trg_TA_pipeline}}
\end{figure*}

Figure \ref{src_trg_TA_pipeline} (right) shows the pipeline that we followed to build the mixture of the outputs using both seeds. Although the data obtained from $TA_{trg}$ are not always useful for adapting an MT model for the test set, mixing the data selected using the test set and the approximated target side can lead to improvements \citep{poncelas2018data}.


\section{Fine-tuning Models with Synthetic Data}

The work of \citet{sennrich2015improving} showed that NMT models can be improved by adding synthetic training data. In their work they use monolingual sentences in the target language and translate them into the source-language with an NMT model. This creates a parallel corpus in which the source side has been artificially generated and the target side is human-produced data (and hence, the fluency of the translation will not be affected). Models built with back-translated data alone (or mixed with back-translated data) can have a performance comparable to those built with real data \citep{poncelas2018investigating}.

In this work we want to explore the performance of NMT models when fine-tuned with TA-selected synthetic data so they are adapted to a given test set. We are interested in exploring three main Research Questions (RQ):

\begin{itemize}
    \item RQ1: \textbf{Does a model adapted with TA-selected back-translated data achieve improvements over the non-adapted model?}

    The strength of performing the fine-tuning technique is to adapt a model with data in the same domain as the document to be translated. Although TA can retrieve relevant data, we do not know the performance when executed using synthetic data. The artificially-generated sentences may contain unusual {\em n}-grams, so the overlap with the test set is lower. This prevents TA from retrieving relevant sentences.

    \item RQ2: \textbf{Does a model adapted with TA-selected back-translated data perform better than a model adapted with TA-selected authentic data?}
    
    Suppose that using synthetic data for adaptation leads to improvements, we also want to compare the performance to that of a model adapted with TA-retrieved authentic data. The quality of the back-translated (source) data, in terms of being an exact translation of the target, is expected to be lower than that of the source-side in the corresponding authentic sentence pairs (which were after all created by human translators). However, the authentic data have already been used to build the model to be adapted, whereas the selected artificial (source) sentences is a set of newly generated data, which may add useful new information not present in the original authentic data set. For this reason, the selected synthetic data might add more value to training the model and may also improve generalization. Therefore, fine-tuning with selected back-translated data may yield larger performance gains than fine-tuning with (repeated) authentic sentences.
    
    \item RQ3: \textbf{Is it preferable to follow the batch or the online processing?}

    As both processing (batch and online) retrieve different subsets of data, we want to study the performance of the models when they are adapted with a mixture of both outputs. The strategy we follow to combine the outputs is to concatenate them in different proportion in a similar way (using different sizes of $\alpha$) as explained in Section \ref{sec:approximated_target_side}.
\end{itemize}



\section{Experiments}

\subsection{Experimental settings}

We build German-to-English models with the parallel data provided in the WMT 2015 \citep{bojar-EtAl:2015:WMT} (\textit{training data}). All data sets are tokenized and truecased. We also apply Byte Pair Encoding (BPE) \citep{sennrich2016neural} with 89500 merge operations. The synthetic data are built by translating the target-side (English) into the source language (German). We use an NMT model built with 1M randomly-selected sentences.

The NMT models are built using OpenNMT-py \footnote{\url{https://github.com/OpenNMT/OpenNMT-py}} \citep{opennmt} with the default parameter values: 2-layer LSTM with 500 hidden units, vocabulary size of 50000 words for each language.

All the models built are evaluated on two test sets using BLEU \citep{papineni2002bleu}, TER \citep{snover2006study} and METEOR \citep{banerjee2005meteor} evaluation metrics. These metrics provide an estimation of the quality of the translation compared to a human-translated reference. The two test sets used to evaluate the models are: (i) \textit{NEWS test set} provided in WMT 2015 News Translation Task; and (ii) \textit{BIO test set}, the Cochrane \footnote{\url{http://www.himl.eu/test-sets}} dataset from the WMT 2017 biomedical translation shared task \citep{yepes2017findings}.

In each table, we mark in bold the scores that are better than the baseline, and if they constitute a statistically significant improvement (at level p=0.01) we mark them with an asterisk. This was computed with multeval \citep{clark2011better} using bootstrap resampling \citep{koehn04}.

\subsection{Model Adaptation with Subsets of Data}

The general-domain model used in this work as baseline is an NMT model trained with the complete training dataset for 13 epochs. The result of the model can be seen in Table \ref{table:baseline_13}


\begin{table}[!htbp]
\centering
\caption{ 
Results of the general-domain model evaluated in the NEWS test set and BIO test set.
\label{table:baseline_13}}
\begin{center}
\begin{tabular}{ |p{2cm}|p{2cm}|p{2cm}|}
\hline
	&	NEWS	&	BIO	\\
\hline	
BLEU	&	0.2634	&	0.3314	\\
TER 	&	0.5441	&	0.4679	\\
METEOR	&	0.3009	&	0.3457	\\
\hline	
\end{tabular}
\end{center}
\end{table}

The experiments carried out consist of using INR and FDA to select different sizes of data: 100K, 200K and 500K sentence pairs. In INR method, a low value of $t$ causes the method to be more strict and retrieve less sentences. We use the larger value so the execution does not exceed 48 hours (i.e. $t=80$ for NEWS test set and $t=640$ for BIO test set). However, the amount of sentences retrieved are below 500K, so in the experiments we only evaluate the models adapted with 100K and 200K INR-selected sentences. The sentences retrieved are used to adapt the general-domain model. In particular, we adapt the 12th epoch of the model by fine-tuning it with the selected data.

\begin{table*}[!htbp]
\centering
\caption{Results of the models built with different sizes of $INR_{src}$ and $INR_{trg}$ using authentic data.}
\begin{center}
\begin{tabular}{ |p{0.5cm}|p{1.5cm}|p{1.5cm}|p{1.5cm}|p{1.5cm}|p{1.5cm}|p{1.5cm}|p{1.5cm}|}
\hline
 &&baseline&	$\alpha=1$  & $\alpha=0.75$ & $\alpha=0.50$ & $\alpha=0.25$ & $\alpha=0$ \\
\hline	
\multicolumn{8}{|c|}{NEWS}\\
\hline										
\multirow{3}{*}{\rotatebox[origin=c]{90}{\centering 100K }}
&BLEU	&	0.2634	&	\bf0.2649	&	\bf0.2659	&	\bf0.2664*	&	\bf0.2655	&	\bf0.2659*	\\
&TER	&	0.5441	&	\bf0.5419	&	\bf0.5408*	&	\bf0.5417*	&	\bf0.5413	&	\bf0.5430*	\\
&METEOR	&	0.3009	&	\bf0.3021*	&	\bf0.3030*	&	\bf0.3037*	&	\bf0.3033*	&	\bf0.3034*	\\
\hline													
\multirow{3}{*}{\rotatebox[origin=c]{90}{\centering 200K}}	
&BLEU	&	0.2634	&	\bf0.2644	&	\bf0.2661*	&	\bf0.2666*	&	\bf0.2655	&	\bf0.2649	\\
&TER	&	0.5441	&	\bf0.5435	&	\bf0.5410*	&	\bf0.5406*	&	\bf0.5413*	&	\bf0.5437*	\\
&METEOR	&	0.3009	&	\bf0.3012	&	\bf0.3025*	&	\bf0.3028*	&	\bf0.3029*	&	\bf0.3027*	\\
\hline													
\multicolumn{8}{|c|}{BIO}\\													
\hline													
\multirow{3}{*}{\rotatebox[origin=c]{90}{\centering 100K}}		
&BLEU	&	0.3314	&	\bf0.3352*	&	\bf0.3346	&	\bf0.3347	&	\bf0.3370*	&	\bf0.3339	\\
&TER	&	0.4679	&	\bf0.4592*	&	\bf0.4631	&	\bf0.462	&	\bf0.4591*	&	\bf0.4605*	\\
&METEOR	&	0.3457	&	\bf0.3477	&	\bf0.3478	&	\bf0.3463	&	\bf0.3488*	&	\bf0.3475	\\
\hline													
\multirow{3}{*}{\rotatebox[origin=c]{90}{\centering 200K}}								
&BLEU	&	0.3314	&	\bf0.3388*	&	\bf0.3362*	&	\bf0.3403*	&	\bf0.3386*	&	\bf0.3343	\\
&TER	&	0.4679	&	\bf0.459*	&	\bf0.4589*	&	\bf0.457*	&	\bf0.4563*	&	\bf0.4590*	\\
&METEOR	&	0.3457	&	\bf0.3494*	&	\bf0.3477	&	\bf0.3502*	&	\bf0.3489*	&	\bf0.3495*	\\
\hline	
\end{tabular}
\label{table:results_srctrag_INR_auth}
\end{center}
\end{table*}

\begin{table*}[!htbp]
\centering
\caption{Results of the models built with different sizes of $FDA_{src}$ and $FDA_{trg}$ using authentic data. }
\begin{center}
\begin{tabular}{ |p{0.5cm}|p{1.5cm}|p{1.5cm}|p{1.5cm}|p{1.5cm}|p{1.5cm}|p{1.5cm}|p{1.5cm}|}
\hline
 &&baseline&	$\alpha=1$  & $\alpha=0.75$ & $\alpha=0.50$ & $\alpha=0.25$ & $\alpha=0$ \\
\hline	
\multicolumn{8}{|c|}{ NEWS}\\
\hline	
\multirow{3}{*}{\rotatebox[origin=c]{90}{\centering 100K}}
&BLEU	&	0.2634	&	\bf0.2649	&	\bf0.2665*	&	\bf0.2642*	&	\bf0.2643	&	0.2633	\\
&TER	&	0.5441	&	\bf0.5421	&	\bf0.5412*	&	\bf0.5413*	&	\bf0.5416*	&	\bf0.5416*	\\
&METEOR	&	0.3009	&	\bf0.3021*	&	\bf0.3027*	&	\bf0.3022*	&	\bf0.3019	&	\bf0.3020	\\
\hline													
\multirow{3}{*}{\rotatebox[origin=c]{90}{\centering 200K}}													
&BLEU	&	0.2634	&	\bf0.2655	&	\bf0.2665*	&	\bf0.2651	&	\bf0.2652	&	\bf0.2654*	\\
&TER	&	0.5441	&	\bf0.5417*	&	\bf0.5412*	&	\bf0.5413*	&	\bf0.5421*	&	\bf0.5404*	\\
&METEOR	&	0.3009	&	\bf0.3024*	&	\bf0.3027*	&	\bf0.3025*	&	\bf0.3025*	&	\bf0.3027*	\\
\hline													
\multirow{3}{*}{\rotatebox[origin=c]{90}{\centering 500K}}													
&BLEU	&	0.2634	&	\bf0.264*	&	\bf0.2658*	&	\bf0.2671*	&	\bf0.2654	&	\bf0.2650	\\
&TER	&	0.5441	&	0.5447	&	\bf0.5414*	&	\bf0.5412*	&	\bf0.5415*	&	\bf0.5404*	\\
&METEOR	&	0.3009	&	\bf0.3010*	&	\bf0.3028*	&	\bf0.3028*	&	\bf0.3024*	&	\bf0.3028*	\\
\hline													
\multicolumn{8}{|c|}{BIO}\\													
\hline													
\multirow{3}{*}{\rotatebox[origin=c]{90}{\centering 100K}}													
&BLEU	&	0.3314	&	\bf0.3368*	&	\bf0.3377*	&	\bf0.3391*	&	\bf0.339*	&	\bf0.3331	\\
&TER	&	0.4679	&	\bf0.4597*	&	\bf0.4611*	&	\bf0.4599*	&	\bf0.4597*	&	\bf0.4649	\\
&METEOR	&	0.3457	&	\bf0.3471	&	\bf0.3473	&	\bf0.3476	&	\bf0.3485	&	\bf0.3463	\\
\hline													
\multirow{3}{*}{\rotatebox[origin=c]{90}{\centering 200K}}													
&BLEU	&	0.3314	&	\bf0.3396*	&	0.3414*	&	\bf0.3375*	&	\bf0.3391*	&	\bf0.3370*	\\
&TER	&	0.4679	&	\bf0.4564*	&	\bf0.459*	&	\bf0.4574*	&	\bf0.4596*	&	\bf0.4572*	\\
&METEOR	&	0.3457	&	\bf0.3501*	&	\bf0.3503*	&	\bf0.3491*	&	\bf0.3484*	&	\bf0.3496*	\\
\hline													
\multirow{3}{*}{\rotatebox[origin=c]{90}{\centering 500K}}													
&BLEU	&	0.3314	&	\bf0.3375*	&	\bf0.3406*	&	\bf0.3358*	&	\bf0.3354*	&	\bf0.3336	\\
&TER	&	0.4679	&	\bf0.4592*	&	\bf0.4552*	&	\bf0.4593*	&	\bf0.4574*	&	\bf0.4617	\\
&METEOR	&	0.3457	&	\bf0.3492*	&	\bf0.3496*	&	\bf0.3485	&	\bf0.3494*	&	\bf0.3485*	\\
\hline	
\end{tabular}
\label{table:results_srctrag_FDA_auth}
\end{center}
\end{table*}

In Table \ref{table:results_srctrag_INR_auth} and Table \ref{table:results_srctrag_FDA_auth} we show the performance of the models when fine-tuned with different sizes of selected authentic data. In the tables we also indicate the proportions of data selected using the test set or the approximated target side as seed.

As we can see, the performance of the adapted models are higher than that of the general-domain model (Table \ref{table:baseline_13}). In addition, using a mixture of $TA_{src}$ and $TA_{trg}$ (columns $\alpha=0.75$, $\alpha=0.50$ and $\alpha=0.25$) can achieve a higher performance than $TA_{src}$ or $TA_{trg}$ alone.


In our experiments we follow the same procedure using synthetic data in order to perform comparisons among the general-domain model, models adapted with authentic data, and models adapted with synthetic data.

\FloatBarrier
\section{Results}

The results of the models adapted with synthetic data are shown in Table \ref{table:results_srctrag_INR_synth} (INR method) and Table \ref{table:results_srctrag_FDA_synth} (FDA method). In order to answer RQ1, we include in the first column, as baseline, the performance of the 13th epoch of the general-domain model (Table \ref{table:baseline_13}). We mark in bold those scores that indicate a better performance than the baseline and add an asterisk if they are statistically significant at level p=0.01.

In the tables we observe that adapted models with artificial data tend to perform better on NEWS test set than BIO test set (e.g. BLEU scores are only higher in the NEWS test set). This manifests that the domain of the model used for back-translating plays an important role. In our experiments the above model is closer to the news domain because it was built using a sample of the authentic training data.

METEOR scores of adapted models are higher than those of the general-domain model for both test sets, and in many cases the improvements are statistical significant (with p=0.001). In contrast, TER scores are lower than the baseline. This may be caused by the synonym or conjugation chosen by the adapted model. For example, the sentence \say{auch Schulen} is translated by the general-domain model as \say{schools too} (the same as in the reference), but adapted model produced \say{also schools}.




\begin{table*}[!htbp]
\centering
\caption{Results of the models built with different sizes of $INR_{src}$ and $INR_{trg}$ using back-translated data.}
\begin{center}
\begin{tabular}{ |p{0.5cm}|p{1.5cm}|p{1.5cm}|p{1.5cm}|p{1.5cm}|p{1.5cm}|p{1.5cm}|p{1.5cm}|}
\hline
 &&	baseline  &	$\alpha=1$  & $\alpha=0.75$ & $\alpha=0.50$ & $\alpha=0.25$ & $\alpha=0$ \\
\hline	
\multicolumn{8}{|c|}{NEWS}\\
\hline	
\multirow{3}{*}{\rotatebox[origin=c]{90}{\centering 100K}}
&BLEU	&	0.2634	&	\bf0.2664	&	\bf0.267	&	\bf0.2671	&	\bf0.2679*	&	\bf0.2675*	\\	
&TER	&	0.5441	&	0.5492	&	0.5496	&	0.55	&	0.5496	&	0.5513	\\	
&METEOR	&	0.3009	&	\bf0.3058*	&	\bf0.3062*	&	\bf0.3063*	&	\bf0.3067*	&	\bf0.3061*	\\	
\hline	
\multirow{3}{*}{\rotatebox[origin=c]{90}{\centering 200K}}	
&BLEU	&	0.2634	&	\bf0.2666	&	\bf0.2673*	&	\bf0.2678*	&	\bf0.2673*	&	\bf0.2672*    \\	
&TER	&	0.5441	&	0.5485	&	0.5486	&	0.5478	&	0.5481	&	0.5481    \\	
&METEOR	&	0.3009	&	\bf0.3064*	&	\bf0.3061*	&	\bf0.3068*	&	\bf0.3066*	&	\bf0.3068*    \\	
\hline	
\multicolumn{8}{|c|}{BIO}\\	
\hline	
\multirow{3}{*}{\rotatebox[origin=c]{90}{\centering 100K}}	
&BLEU	&	0.3314	&	0.324	&	0.327	&	0.3263	&    0.3269	&	0.3251	\\	
&TER	&	0.4679	&	0.4762	&	0.4747	&	0.4753	&	0.4751	&	0.4764	\\	
&METEOR	&	0.3457	&	\bf0.3486	&	\bf0.3490	&	\bf0.3502*	&	\bf0.351*	&	\bf0.3489	\\	
\hline	
\multirow{3}{*}{\rotatebox[origin=c]{90}{\centering 200K}}	
&BLEU	&	0.3314	&	0.3241	&	0.3255	&	0.3255	&	0.3254	&	0.3251	\\
&TER	&	0.4679	&	0.4782	&	0.4755	&	0.4732	&	0.4742	&	0.4745	\\
&METEOR	&	0.3457	&	\bf0.3487	&	\bf0.3501*	&	\bf0.3508*	&	\bf0.3509*	&	\bf0.3505*	\\
\hline	
\end{tabular}
\label{table:results_srctrag_INR_synth}
\end{center}
\end{table*}

\begin{table*}[!htbp]
\centering
\caption{Results of the models built with different sizes of $FDA_{src}$ and $FDA_{trg}$ using back-translated data.}
\begin{center}
\begin{tabular}{ |p{0.5cm}|p{1.5cm}|p{1.5cm}|p{1.5cm}|p{1.5cm}|p{1.5cm}|p{1.5cm}|p{1.5cm}|}
\hline
 &&	baseline &	$\alpha=1$  & $\alpha=0.75$ & $\alpha=0.50$ & $\alpha=0.25$ & $\alpha=0$ \\
\hline														
\multicolumn{8}{|c|}{NEWS}\\														
\hline
\multirow{3}{*}{\rotatebox[origin=c]{90}{\centering 100K}}	
&BLEU	&	0.2634	&	\bf0.2639	&	\bf0.2654	&	\bf0.264	&\bf0.2655	&	\bf0.2672*		\\	
&TER	&	0.5441	&	0.5525	&	0.5509	&	0.5522	&	0.5511	&	0.5493		\\
&METEOR	&	0.3009	&	\bf0.305*	&	\bf0.3054*	&	\bf0.3051*	&	\bf0.3055*	&	\bf0.3062*		\\
\hline														
\multirow{3}{*}{\rotatebox[origin=c]{90}{\centering 200K}}														
&BLEU	&	0.2634	&	\bf0.2655	&	\bf0.2658	&	\bf0.2663	&	\bf0.2666	&	\bf0.2679*		\\
&TER	&	0.5441	&	0.5497	&	0.5512	&	0.5504	&	0.5493	&	0.5484		\\
&METEOR	&	0.3009	&	\bf0.3051*	&	\bf0.3053*	&	\bf0.306*	&	\bf0.3055*	&	\bf0.3063*		\\
\hline														
\multirow{3}{*}{\rotatebox[origin=c]{90}{\centering 500K}}														
&BLEU	&	0.2634	&	\bf0.2662	&	\bf0.2674*	&	\bf0.2668	&	\bf0.2679*	&	\bf0.2664		\\
&TER	&	0.5441	&	0.5483	&	0.5494	&	0.5501	&	0.5488	&	0.5489		\\
&METEOR	&	0.3009	&	\bf0.3061*	&	\bf0.3068*	&	\bf0.3062*	&	\bf0.3068*	&	\bf0.3062*		\\
\hline														
\multicolumn{8}{|c|}{BIO}\\														
\hline														
\multirow{3}{*}{\rotatebox[origin=c]{90}{\centering 100K}}														
&BLEU	&	0.3314	&	0.3228	&	0.3248	&	0.3238	&	0.3254	&	0.3262		\\
&TER	&	0.4679	&	0.4755	&	0.475	&	0.4751	&	0.4742	&	0.4744		\\
&METEOR	&	0.3457	&	\bf0.349	&	\bf0.3488	&	\bf0.3497*	&	\bf0.3521*	&	\bf0.3500*		\\
\hline														
\multirow{3}{*}{\rotatebox[origin=c]{90}{\centering 200K}}														
&BLEU	&	0.3314	&	0.3214	&	0.3245	&	0.3258	&	0.3255	&	0.3241		\\
&TER	&	0.4679	&	0.478	&	0.4743	&	0.4737	&	0.4751	&	0.4749		\\
&METEOR	&	0.3457	&	\bf0.3487	&	\bf0.3495	&	\bf0.3501*	&	\bf0.349	&	\bf0.3482		\\
\hline														
\multirow{3}{*}{\rotatebox[origin=c]{90}{\centering 500K}}														
&BLEU	&	0.3314	&	0.3215	&	0.3223	&	0.3229	&	0.3241	&	0.3226		\\
&TER	&	0.4679	&	0.4842	&	0.4843	&	0.4817	&	0.4813	&	0.4811		\\
&METEOR	&	0.3457	&	\bf0.3478	&	\bf0.3488	&	\bf0.3486	&	\bf0.3491	&	\bf0.349		\\
\hline	
\end{tabular}
\label{table:results_srctrag_FDA_synth}
\end{center}
\end{table*}

\subsection{Model Adaptation with Synthetic Data}

In our experiments, the back-translated data used for the adaptation are new data unseen by the model (the authentic data used to adapt the models presented in tables \ref{table:results_srctrag_INR_auth} and \ref{table:results_srctrag_FDA_auth} are subsets of the same data used to build the general-domain model). The outcomes observed in the experiments show that adapting the models with synthetic data does not achieve as good results as adapting them with authentic data (which answers the RQ2). If we compare cell-wise (i.e. same value of $\alpha$ and same size of selected sentences) tables \ref{table:results_srctrag_INR_auth} and \ref{table:results_srctrag_INR_synth} or tables \ref{table:results_srctrag_FDA_auth} and \ref{table:results_srctrag_FDA_synth} we see slight improvements for the BLEU and METEOR scores for the news test set (NEWS subtables). However, none of these are statistically significant at p=0.01.

As mentioned previously, the sentences produced by the model used for back-translation may contain mistakes such as word-ordering, incorrect translations etc. which reduces the potential sentences that TA can retrieve. For example, in our experiments we find the following sentence in the NEWS test set \say{Auf der Hüpfburg beim Burggartenfest war am Sonnabend einiges los.} (according to the reference \say{Something is happening on the bouncy castle at the Burggartenfest.}) contains the word \say{Hüpfburg} (\say{bouncy castle}) which is used by TA to retrieve sentences. There are 18 occurrences of this word in the authentic data set. However, in the synthetic data there are no instances of this word. Instead, the back-translated counterparts of sentences containing \say{Hüpfburg} include words such as \say{bouncer} (copied from the English side) or \say{bounmit} (a word that does not exist). Nevertheless, in some cases back-translated sentences may be closer to literal translation than those found in the authentic set \citep{poncelas2018data,poncelas2018adapt}. For example, in the authentic data set we find the sentence-pair ⟨\say{er ist verheiratet und hat zwei Kinder.},\say{since then, he has had a long career on stage, in film and on television. he has also established himself as a singer and an author in recent years.}⟩ which do not convey the same meaning. However, the machine-produced source-side is \say{seitdem hat er eine lange Karriere auf der Bühne, im Film und im Fernsehen absolviert und hat sich auch als Sängerin und Autor in den letzten Jahren etabliert} which is closer in meaning to the target-side sentence. Another example is the pair ⟨\say{10 \%!},\say{one tenth!}⟩. Although, they have the same meaning, in the back-translated counterpart the source-side sentence is \say{ein Zehntel!}, which is a literal translation.



\subsection{Batch and Online Processing}

In order to answer RQ3 we need to compare columns $\alpha=1$ (batch processing, i.e. extract from back-translated data using the test set) and $\alpha=0$ (online processing, i.e. extract from authentic data using the approximated set). In Table \ref{table:results_srctrag_INR_synth} and Table \ref{table:results_srctrag_FDA_synth} we see that in our experiments following the online process the results tend to be better.

Using an approximated target side as seed is risky, as it can be of low quality. For example, the sentence \say{Das Buch wurde neu für 48\$ verkauft.} (\say{The book was selling for \$48 new.}) is translated as \say{The book was sold for 48\$.} by the general-domain model. As we can see, the word \say{new} is omitted in the translation. This means that the TA will not consider the word \say{new} when selecting sentences. 


Despite that, we find that the generated target-side seed may contain {\em n}-grams that better represent the context of the input document. For example, the sentence in the test set \say{Ich liebe es, in einem Probenraum zu sein.} is translated, according to the reference, as \say{I love being in a rehearsal room.}. The model adapted with 100K sentences from $FDA_{src}$ ($\alpha=1$) generates the translation \say{I love to be in a sample room.}, whereas the model adapted with $FDA_{trg}$ ($\alpha=0$) produces a sentence that conveys the same meaning to the reference: \say{I love to be in a rehearsal room.}.

We observe that the occurrences of \say{Proben} (due to BPE, the word is splitted as \say{Proben@@ raum}) are translated as \say{sample} or \say{rehersal} depending on the context. The fact that in the approximated target side the word has been accurately translated as \say{rehearsal room} induces $FDA_{trg}$ to select more sentences that include the term \say{rehearsal}. In contrast, $FDA_{src}$ retrieves sentences based on the word \say{Proben} in the seed (as it is present in the test set). However, in the training data this word has been artificially produced and it replaces words such as \say{Messwasser} (\say{water sample}) or \say{Musterproduktion} (\say{sample production}).

\section{Conclusion and Future Work}

In this paper we have analyzed various use-cases of synthetic data for adapting a general-domain model. We have seen that using a TA it is possible to obtain sentences from synthetic data that can improve the model, even if the sentences used for adaptation are an artificial version of the same sentences used to construct the general model.

In addition, we have seen that performing the adaptation online, extracting just the necessary monolingual target-language sentences (using an approximated translation of the test set as seed) and back-translating them afterwards, is a reasonable approach that can even perform better than selecting directly from synthetic sentences.

In the future, we want to further extend this research and explore the effects on the performance of combining both authentic and synthetic data or the use of forward-translation \citep{chinea2017adapting}. In addition, we are interested in exploring whether the results observed in this paper are the same when using other language pairs or other configurations of INR and FDA \citep{poncelasextending,poncelas2017applying}.

\section*{Acknowledgements}
This research has been supported by the ADAPT Centre for Digital Content Technology which is funded under the SFI Research Centres Programme (Grant 13/RC/2106) and is co-funded under the European Regional Development Fund.

\noindent 
\includegraphics[width=1cm]{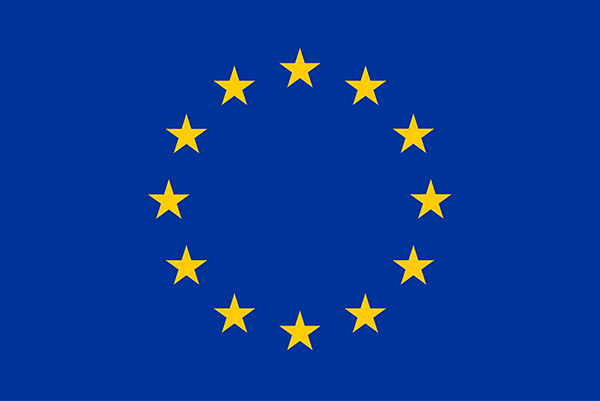}
This work has also received funding from the European Union’s Horizon 2020 research and innovation programme under the Marie Skłodowska-Curie grant agreement No 713567.

\FloatBarrier

\bibliographystyle{splncs}
\bibliography{main.bib}

\end{document}